\documentclass[letterpaper]{spie}

\usepackage{pslatex} 

\usepackage{graphicx}
\graphicspath{{./figures/}}
\usepackage{subcaption}

\usepackage{paralist}
\renewenvironment{enumerate}{\begin{compactenum}}{\end{compactenum}}

\usepackage{amsmath}
\usepackage{amsopn}
\usepackage{amssymb}
\usepackage{mathtools}

\DeclareMathOperator*{\argmin}{arg\,min}

\DeclareMathOperator{\Id}{Id}

\DeclareMathOperator*{\unvectmo}{unvec}
\newcommand\unvect[1]{\unvectmo({#1})}
\DeclareMathOperator*{\vectmo}{vec}
\newcommand\vect[1]{\vectmo({#1})}
\DeclareMathOperator*{\vechmo}{vech}
\newcommand\vech[1]{\vechmo({#1})}
\DeclareMathOperator*{\diagmo}{diag}
\newcommand\diag[1]{\diagmo({#1})}

\DeclareMathOperator*{\tvmo}{TV}
\newcommand\tv[1]{\tvmo({#1})}
\DeclareMathOperator*{\tvhmo}{TV^H}

\DeclareMathOperator{\ad}{ad}
\DeclareMathOperator{\sinch}{sinch}

\usepackage{hyperref}
\title{A Log-Euclidean and Total Variation based Variational Framework for Computational Sonography}

\author{Jyotirmoy Banerjee\supit{a},
Premal A. Patel\supit{a},
Fred Ushakov\supit{b}, 
Donald Peebles\supit{b},
Jan Deprest\supit{a,c},\\
S\'ebastien Ourselin\supit{a},
David Hawkes\supit{a} and
Tom Vercauteren\supit{a,c}
\skiplinehalf
\supit{a}Wellcome / EPSRC Centre for Interventional and Surgical
Sciences, UCL, UK; \\
\supit{b}Fetal Medicine Unit, University College London Hospital,
UK; \\
\supit{c} Organ Systems Unit, Katholieke Universiteit Leuven, Belgium
}

\begin{document}

\maketitle

\begin{abstract}
We propose a spatial compounding technique and variational framework
to improve 3D
ultrasound image quality by compositing multiple ultrasound volumes
acquired from different probe orientations. In the composite volume,
instead of intensity values, we estimate a tensor at every
voxel. The resultant tensor image encapsulates the directional
information of the underlying imaging data and can be used to generate
ultrasound volumes from arbitrary, potentially unseen, probe positions. Extending the
work of Hennersperger et al.\cite{DBLP:conf/miccai/HennerspergerBM15},
we introduce a log-Euclidean framework to ensure that the tensors
are positive-definite, eventually ensuring non-negative images. 
Additionally, we regularise the underpinning ill-posed variational problem while
preserving edge information by relying on a total variation
penalisation of the tensor field in the log domain.
We present results
on in vivo human data to show the
efficacy of the approach. 
\end{abstract}


\section{Introduction}\label{sec:Introduction}
Ultrasound (US) probes used in diagnostic medicine emit sound in the
frequency range 1 to 20 MHz and receive echoes reflected back from the
tissues being imaged. The strength of the signal and the time taken to
return back to the probe is used to produce the images. The degree of
sound reflection depends on surface structure and angle between
tissue surface and US beam. The position and
orientation of the probe therefore plays an important role in the
appearance of US
images. The complex physics of US image
formation makes images highly direction dependent.
It is therefore customary for the ultrasonographer to capture a
variety of images of the same object by translating the ultrasound
probe over the body surface.
From an image computing perspective, the resulting redundancy of
images calls for a method to compound the data into a single model of
the imaged object.  

There is a wide variety of existing approaches to ultrasound spatial
compounding.
Statistical approaches (like averaging and median) have been
applied over all the voxels across the source images to generate the
composite volume\cite{DBLP:conf/miccai/WachingerWN07}. Optimal stitching seam has been proposed to 
merge overlapping 3D ultrasound volumes. For instance, Kutarnia et
al.\cite{DBLP:conf/miip/KutarniaP13} treats seam selection as a
voxel labeling problem where each 
label corresponds to one source volume.
The optimal labeling, which defines the seams, is solved using
graphcut so as to minimize the intensity
and gradient difference between
adjacent volume selections.
Recently, computational
sonography\cite{DBLP:conf/miccai/HennerspergerBM15} was suggested
to provide a richer signal representation based on the reconstruction
of tensor fields that 
preserves the directionality components of the anatomy-specific and
direction-depend source images, as opposed to traditional intensity
volume reconstruction. 

Our contributions, in this work,
are the following. First, given a set of previously aligned ultrasound
volumes, we propose
a novel spatial compounding technique and
log-Euclidean variational framework to generate a
composite image from multiple ultrasound volumes.
Image registration is used to align the input images in a common
reference space.
The reconstructed composite image has a richer
representation than each individual ultrasound images. It
encapsulates,
using a tensor representation,
the directionality component of the signal as captured by the different
probe orientations from the input images.
Similar to the previous work of Hennersperger
et al.\cite{DBLP:conf/miccai/HennerspergerBM15}, the tensor
representation allows to generate images under any arbitrary,
potentially unseen,
direction of the probe. However, unlike the previous work, our tensor
representation is guaranteed to be positive-definite, thereby ensuring
generation of non-negative images. 
Second, we regularise the underpinning ill-posed variational problem
while preserving edge information by relying on a total variation
penalisation of the tensor field in the log domain.
%
%
Finally, we demonstrate the
performance of the method on real dataset. 

The paper is structured as follows. In Section~\ref{sec:Mosaic}, we
present the image alignment steps. In Section~\ref{sec:Preliminaries},
we briefly cover the mathematical background. In
Section~\ref{sec:Log-Euclidean}, we introduced our spatial compounding
approach. Finally, in Section~\ref{sec:Experiments and Discussion}, we
discuss the experiments and results. 

\begin{figure}[htb]
 \centering
  \includegraphics[height=0.35\textwidth]{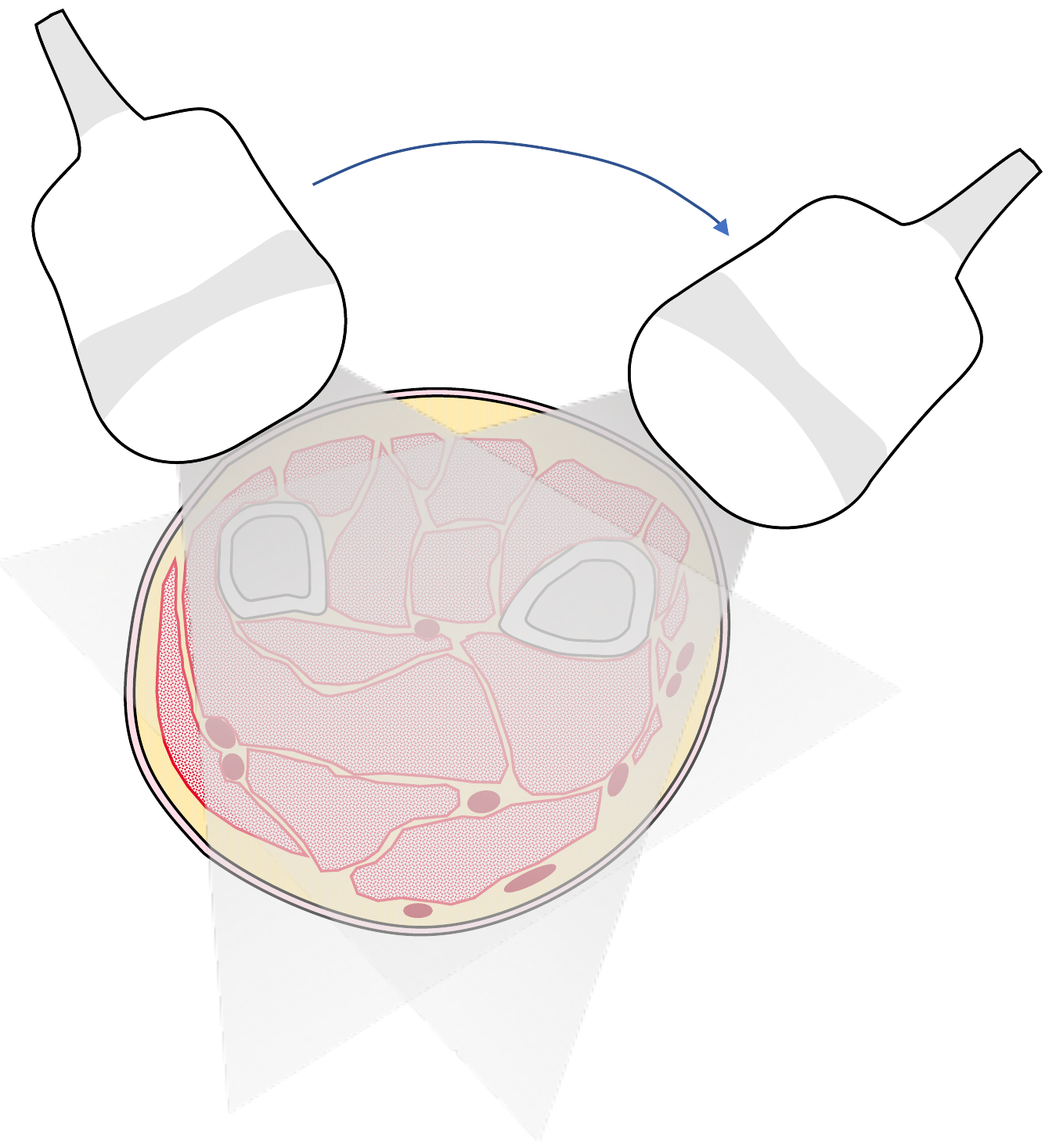} \quad
  \includegraphics[height=0.35\textwidth]{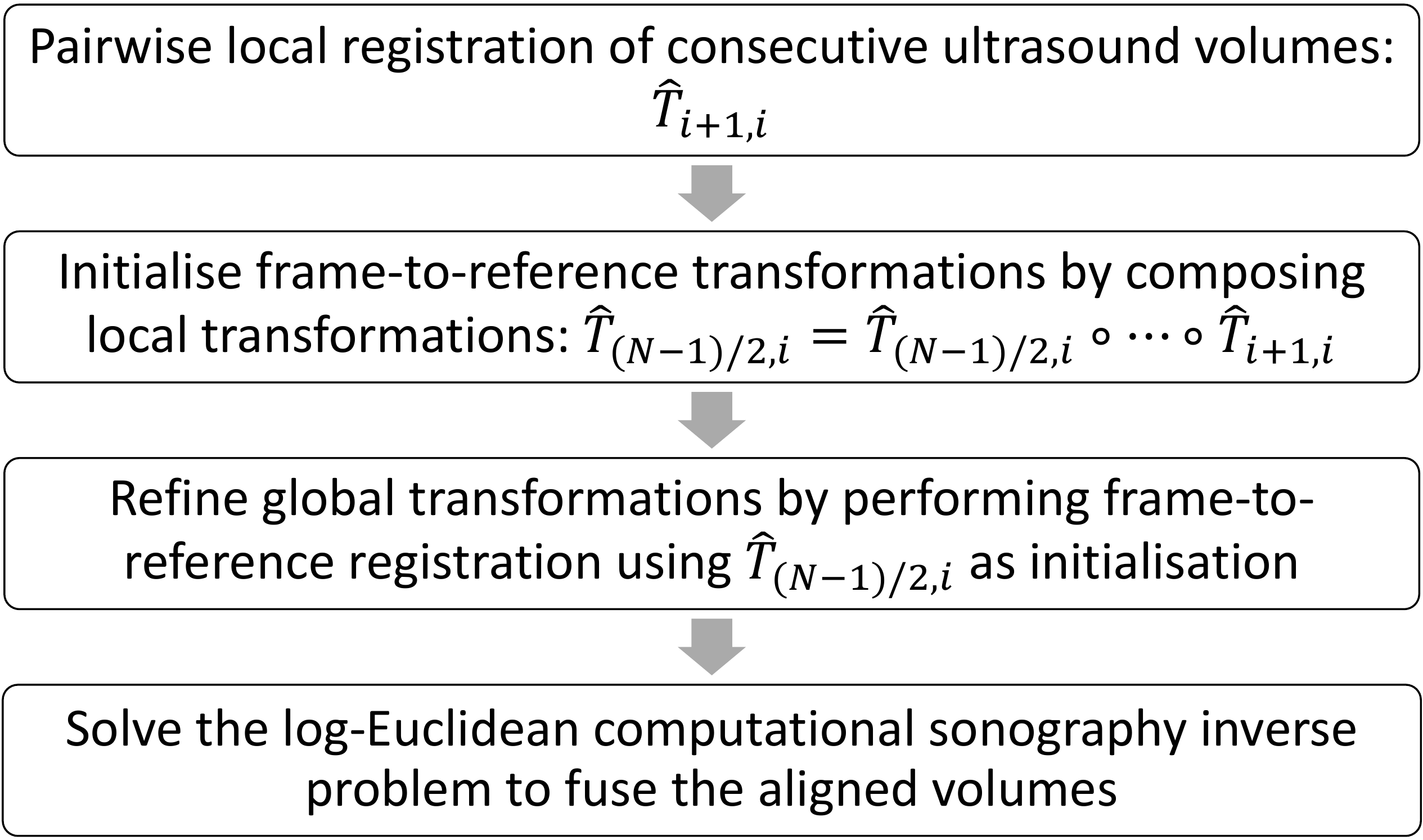}
\caption{Overview of the method. Left: Schematic representation of the
multiview acquisition. Right: Proposed flowchart.}
\label{fig:overview}
\end{figure}

\section{Image Alignment}\label{sec:Mosaic}
As illustrated in Figure~\ref{fig:overview},
given a set of $N$ ultrasound volumes, we register them, using rigid
transformations, to a common reference frame before we apply our spatial
compounding technique. We choose the middle image of the sequence,
i.e.\ the $(N-1)/2$ frame, as the reference image. All the images
are registered to the middle image. However, direct registration
between the $i$th image, where $1 \le i \le N$, to the centre image
may be difficult considering the wide difference in orientation
between the images. To address this we follow the steps
below to register the $i$th image and the $(N-1)/2$th image:
\begin{enumerate}
  \item In the first step, we register neighbouring images using NiftyReg\cite{DBLP:conf/miccai/OurselinRPA00,Modat:JMI:2014}. We obtain
    $\hat{T}_{i+1,i}$ as the transformation from the $i$th image to the $(i+1)$th
    image, where $1 \le i < N$. 
  \item We initialize the transformation $T_{(N-1)/2,i}$ by composing all
    the intermediate transformations calculated in the previous step:
    $\hat{T}_{(N-1)/2,i} = \hat{T}_{(N-1)/2,((N-1)/2)-1} \circ \cdots \circ \hat{T}_{i+2,i+1} \circ \hat{T}_{i+1,i}$.
  \item We refine the transformation $T_{(N-1)/2,i}$ by
    registering, again using NiftyReg\cite{DBLP:conf/miccai/OurselinRPA00,Modat:JMI:2014},
    the $i^{\textrm{th}}$ image and the $(N-1)/2^{\textrm{th}}$ image, starting from the
    transformation $\hat{T}_{(N-1)/2,i}$ from the previous step. 
\end{enumerate}

While this simple approach performed well in the presented experiments, further
work will evaluate more elaborate registration approaches where global
consistency would be achieved by registering all
possible pairs of ultrasound volumes in a
bundle-adjustment fashion\cite{Vercauteren:Media:06} or by relying on
joint registration method\cite{DBLP:conf/cvbia/ZolleiLGW05}.



\section{Mathematical Preliminaries}\label{sec:Preliminaries}

\subsection{Duplication Matrix}\label{sec:Duplication}
Let $A$ be a $k \times k$ matrix.
The operator $\vectmo$ stacks the columns of a matrix into a vector. The
operator $\unvectmo$ does the inverse with $\unvect{\vect{A}} = A$.
If $A$ is symmetric,  $\vectmo(A)$ contains duplicate information.
It is therefore convenient to also consider half-vectorisation,
$\vech{A}$, by eliminating all supra-diagonal elements of $A$.
The duplication operator\cite{Magnus:ET:1986}
$\mathcal{D}:\mathbb{R}^{(k(k+1)/2) \times 1}
\to \mathbb{R}^{(k^2) \times 1}$ duplicates elements of a vector
$U=\vech{A}$, such that $\mathcal{D}(\vech{A}) = \vect{A}$.
In our context where $k=3$, we obtain
\begin{align}
\vech{A} &= (a_{11}, a_{21}, a_{31}, a_{22}, a_{32}, a_{33})^T \\
\vect{A} &= (a_{11}, a_{21}, a_{31}, a_{12}, a_{22}, a_{32}, a_{13},
a_{23}, a_{33})^T
 = \mathcal{D} \cdot \vech{A} \\
\mathcal{D} &=
\begin{bsmallmatrix}
1 & 0 & 0 & 0 & 0 & 0 \\
0 & 1 & 0 & 0 & 0 & 0 \\
0 & 0 & 1 & 0 & 0 & 0 \\
0 & 1 & 0 & 0 & 0 & 0 \\
0 & 0 & 0 & 1 & 0 & 0 \\
0 & 0 & 0 & 0 & 1 & 0 \\
0 & 0 & 1 & 0 & 0 & 0 \\
0 & 0 & 0 & 0 & 1 & 0 \\
0 & 0 & 0 & 0 & 0 & 1 \\
\end{bsmallmatrix}
\end{align}

\subsection{Derivative of the Matrix Exponential}\label{sec:Loewner}
Let $M$ be a diagonalisable matrix
(such as a symmetric matrice).
The derivative of the matrix exponential,
$M \mapsto \exp(M)$,
is provided in
Kalbfleisch et al.\cite{Kalbfleisch:JASA:1985} and
Najfeld et al.\cite{Najfeld:AAM:1995} as:
\begin{align}
\label{eq:exp}
\frac{d\exp(M)}{dM} = (V \otimes V^{-T}) \cdot \diag{\vect{L_{\exp}(\lambda)}}
			       \cdot (V^{-1} \otimes V^{T}) ,
\end{align}
where $\otimes$ is the Kronecker product, $M = V\Lambda V^{-1}$ is the
eigen decomposition of the matrix $M$, $\lambda$ is the vector of
eigenvalues (i.e.\ $\Lambda = \diag{\lambda}$)
and $L_{\exp}(\lambda)$ is the Loewner matrix of the exponential and
the vector of eigenvalues. We have: 
\begin{equation}
L_{\exp}(\lambda) = \frac{\exp(\lambda) \oplus 1 - 1 \oplus
  \exp(\lambda)}{\lambda \oplus 1 - 1 \oplus \lambda},
\end{equation}
where
\begin{equation}
[L_{\exp}(\lambda)]_{ij} =
\left\{
	\begin{array}{ll}
		\exp(\lambda_i)  & \mbox{if } i = j \\
		\frac{(\exp(\lambda_i) - \exp(\lambda_j))}{(\lambda_i - \lambda_j)} & \mbox{if } i \neq j
	\end{array}
\right.
\end{equation}
and where $\oplus$ is the Kronecker sum, i.e.
$P \oplus Q = P \otimes \Id + \Id \otimes Q$.
We note that using a simple Taylor expansion we obtain numerically well-behaved
formulas in case of equal or minor differences between eigen values:
\begin{align}
\frac{\exp(\lambda + \epsilon) - \exp(\lambda)}
   {\lambda + \epsilon - \lambda}
   = \exp(\lambda) ( 1+\frac{\epsilon}{2} + \frac{\epsilon^2}{6} + \dots )
\end{align}
Alternatively, one may also resort to one of the formulas provided by Najfeld
et al.\cite{Najfeld:AAM:1995} for the generic case in which $M$ need
not be differentiable:
\begin{align}
\frac{d \exp(M)}{d M} &= (\Id \otimes \exp(M)) \frac{1 -
  \exp(-\ad_M)}{\ad_M} \\
 &= (\exp(M/2)^T \otimes
\exp(M/2))\sinch(-\ad_{M/2})
\end{align}
with $\ad_M = (- M^T) \oplus M$ providing the adjoint action of a matrix $M$.

\section{Log-Euclidean Computational Sonography}\label{sec:Log-Euclidean}
Given a set of transformed ultrasound volumes, we obtain a composite
volume where each voxel is a $3 \times 3$ tensor by minimizing the
following term as suggested in Hennersperger et
al.\cite{DBLP:conf/miccai/HennerspergerBM15}, 
\begin{equation}
\argmin_{Q_j} \sum_i \sum_j (v_i^T Q_j v_i - I_{ij})^2 ,
\label{eq:old}
\end{equation}
where $Q_j$ is the symmetric tensor at voxel location $j$, $1 \leq j
\leq m$, m is the number of voxels, $1 \leq i \leq n$, $n$ is the
number of images, $v_i$ is the directional vector or probe direction
of the $i$th ultrasound volume and $I_{ij}$ is the voxel intensity.  
As pointed out in Hennersperger et
al.\cite{DBLP:conf/miccai/HennerspergerBM15}, solving the above
equation without specific constraints on $Q_j$
may lead to a non positive definite tensor. 
To ensure a
positive definite tensor $Q_j$, we re-write the above equation using
the log-Euclidean approach of Arsigny et
al.\cite{Arsigny:MRM:06}.
With this approach, positive definite tensors $Q_j$ are parameterised
with arbitrary symmetric matrices $S_j$ through the use of the matrix
exponential $Q_j = \exp(S_j)$.
The operators $\unvectmo$ and $\mathcal{D}$ enables us to write the
symmetric $3 \times 3$ matrices $S_j$ in a parametric form without
redundancies using a vector $X_j \in \mathbb{R}^{6 \times  1}$:
$S_j = \unvect{\mathcal{D}\cdot X_j}$.

We additionally introduce a robust loss function $\rho$ and a total
variation penalisation term to regularise 
the ill-posed problem while preserving edge information. We obtain the
following variational problem:
\begin{equation}
\argmin_{X_j} \sum_i \sum_j
\rho((v_i^T \exp(\unvect{\mathcal{D}\cdot X_j}) v_i - I_{ij})^2)
+ \lambda \cdot \tv{X_j}.
\label{eq:energy}
\end{equation}
A smooth approximation of the total variation regularisation term can be
provided by relying on the Huber loss function, as exemplified in the
1D case below:
\begin{equation}
  \tvhmo(x) = 
    \begin{cases}
      \tfrac{1}{2}(\nabla x)^2 & \text{if } |\nabla x| \leq \delta \\
      \delta(|\nabla x| - \tfrac{1}{2}\delta) & \text{otherwise}
    \end{cases}       
\end{equation}


Equation~\eqref{eq:energy} becomes a non-linear least squares problem that
can efficiently be solved if one can compute the Jacobian of the
residuals.
In this work, we make use of the Levenbeg-Marquardt
algorithm\cite{Madsen:NLLS:2004} available in the Ceres Solver
library\cite{ceres-solver}.
%
%
%
The first term of \eqref{eq:energy} can be rewritten using
\begin{align}
f(X_j) =
\Phi(\exp(\unvect{\mathcal{D}\cdot X_j})),
\end{align}
where
\begin{align}
\Phi(A_j) = v_i^T
A_jv_i - I_{ij}.
\end{align}
Using the chain rule, the Jacobian of $f$ is given as,
\begin{align} 
J_f(X) 
	 = J_{\Phi}( \exp(\unvect{\mathcal{D}\cdot X}) )
         \cdot J_{\exp}(\unvect{\mathcal{D}\cdot X})
	 \cdot J_{\unvectmo}{(\mathcal{D}\cdot X)}
         \cdot J_\mathcal{D}\cdot X 
	 = v_i \otimes v_i \cdot \frac{d\exp(M)}{dM} \cdot 1 \cdot \mathcal{D} .
\end{align}
Combining the terms, the Jacobian of $f$ is given as,
\begin{align} 
J_f(X) = (v_i \otimes v_i) \cdot (V \otimes V^{-T}) \cdot \diag{\vect{L_{\exp}(\lambda)}}
	  \cdot (V^{-1} \otimes V^{T}) \cdot \mathcal{D}.
\end{align}

We highlight that even though the model \eqref{eq:energy} provided
interesting results in our experiment, future work would need to model
more realistically the physics of ultrasound image acquisition
including signal attenuation, and scattering. This could, in the first
instance, be done using an effective but computationally tractable
model such as the one presented by Wein et
al.\cite{DBLP:journals/mia/WeinBKCN08} for CT-Ultrasound registration.

\begin{figure}[htb]
\begin{subfigure}{0.99\textwidth}
  \centering
  \includegraphics[width=0.24\textwidth]{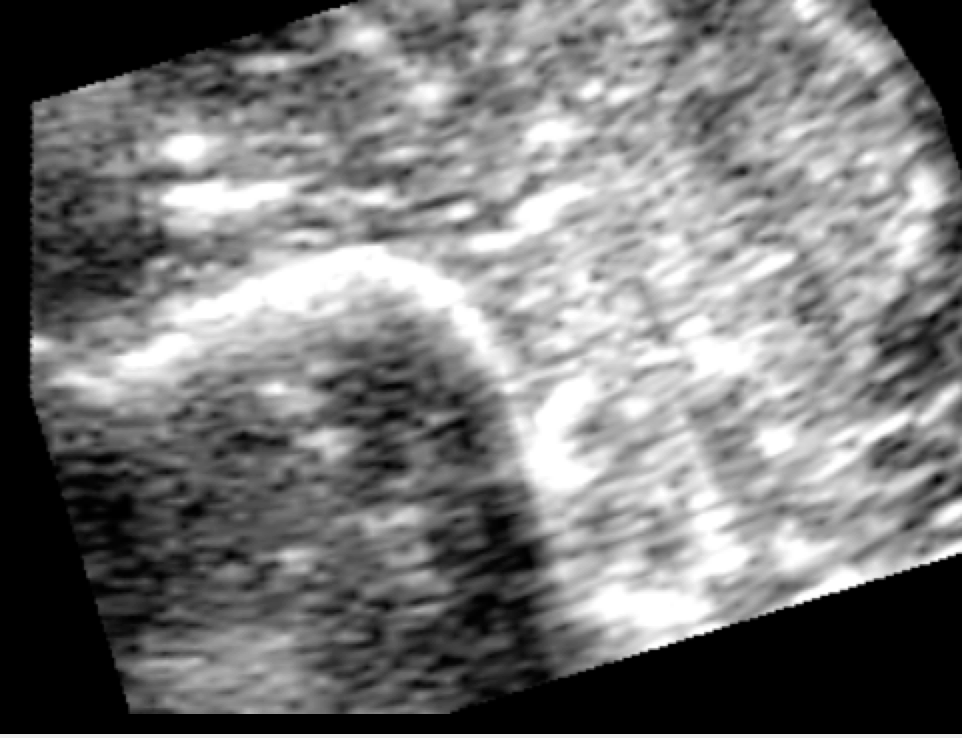}
  \includegraphics[width=0.24\textwidth]{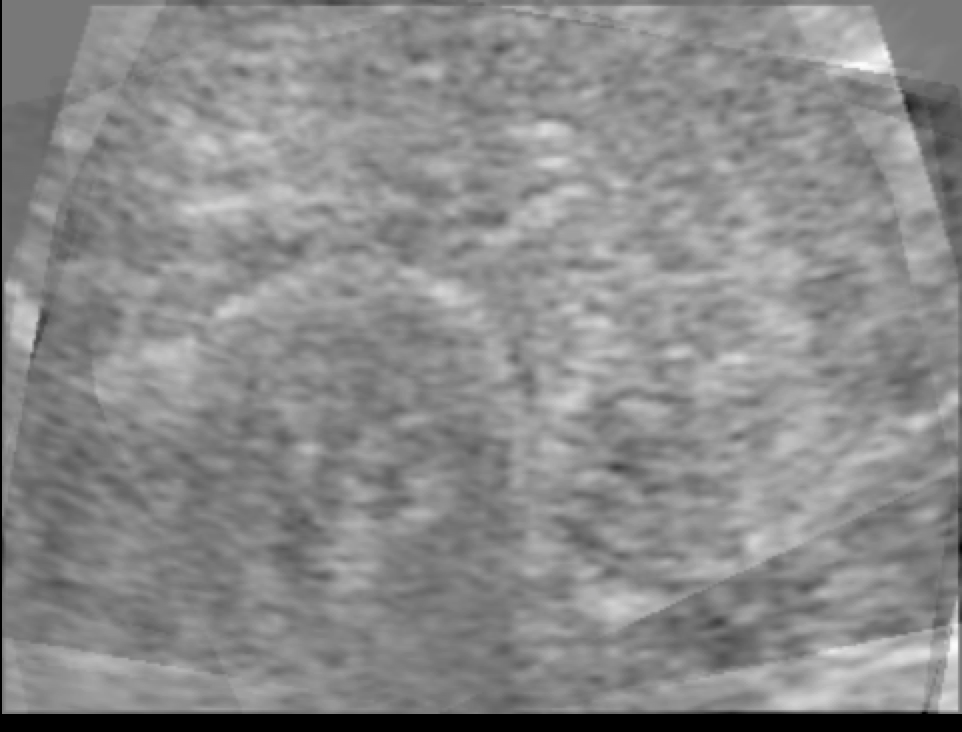}
  \includegraphics[width=0.24\textwidth]{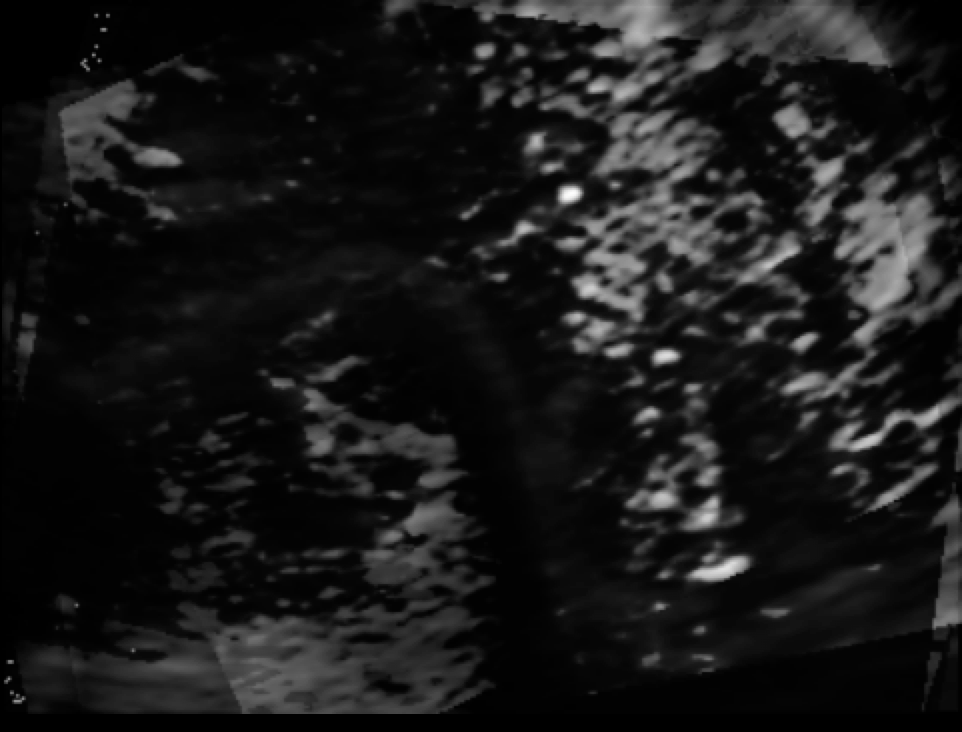}
  \includegraphics[width=0.24\textwidth]{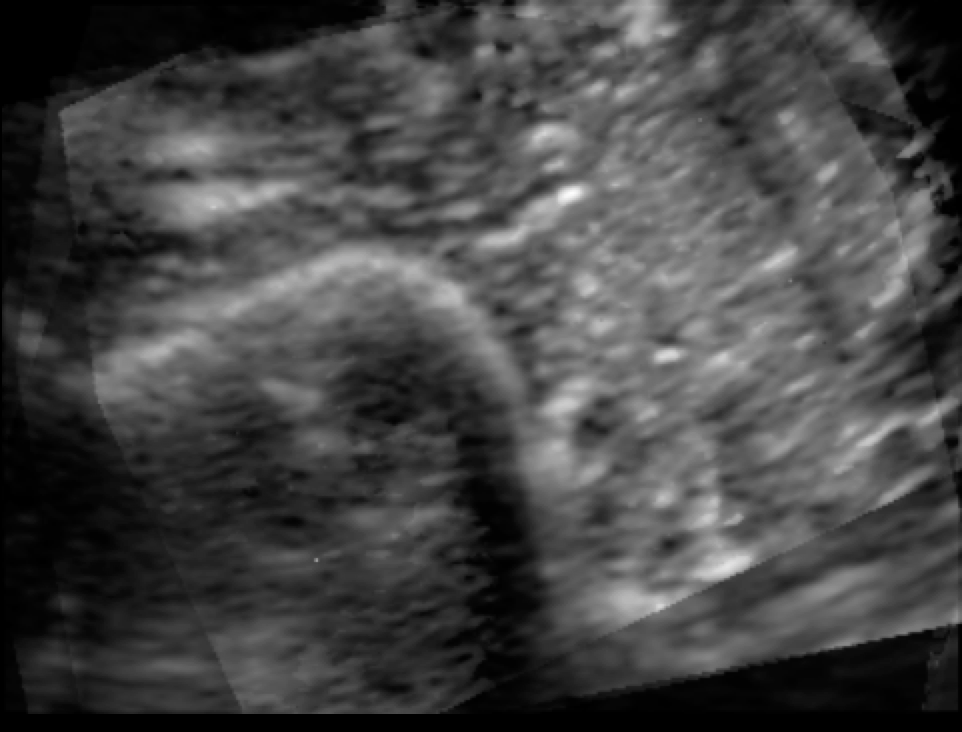}
  \caption{Dataset 1 - 2nd image out of a sequence of nine images}
  \label{fig:sfig12}
\end{subfigure}

\begin{subfigure}{0.99\textwidth}
  \centering
  \includegraphics[width=0.24\textwidth]{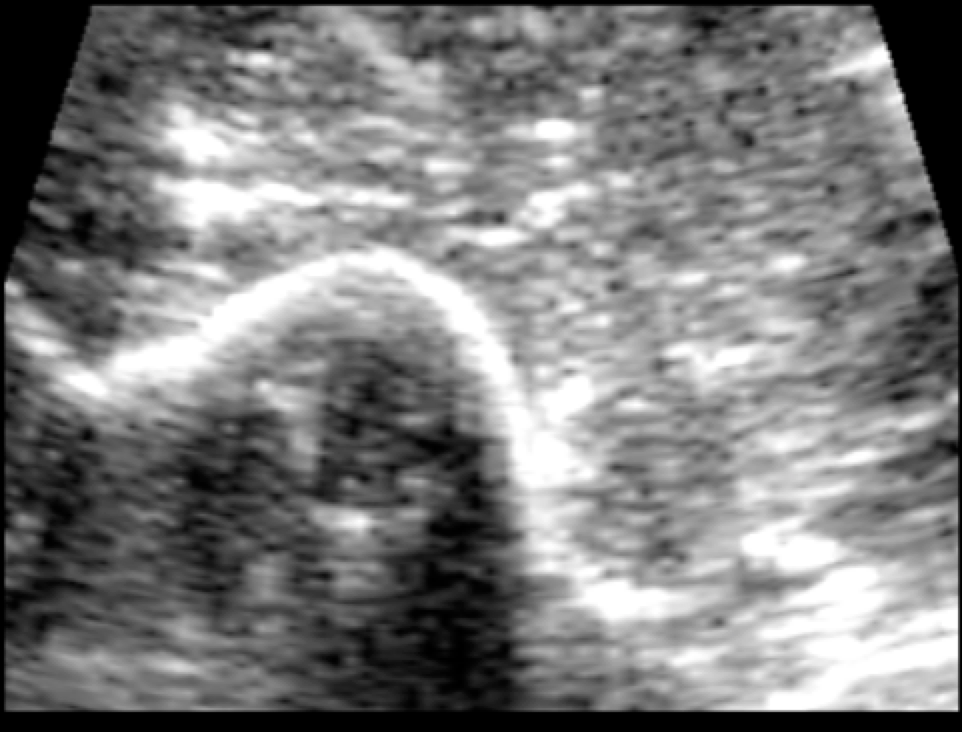}
  \includegraphics[width=0.24\textwidth]{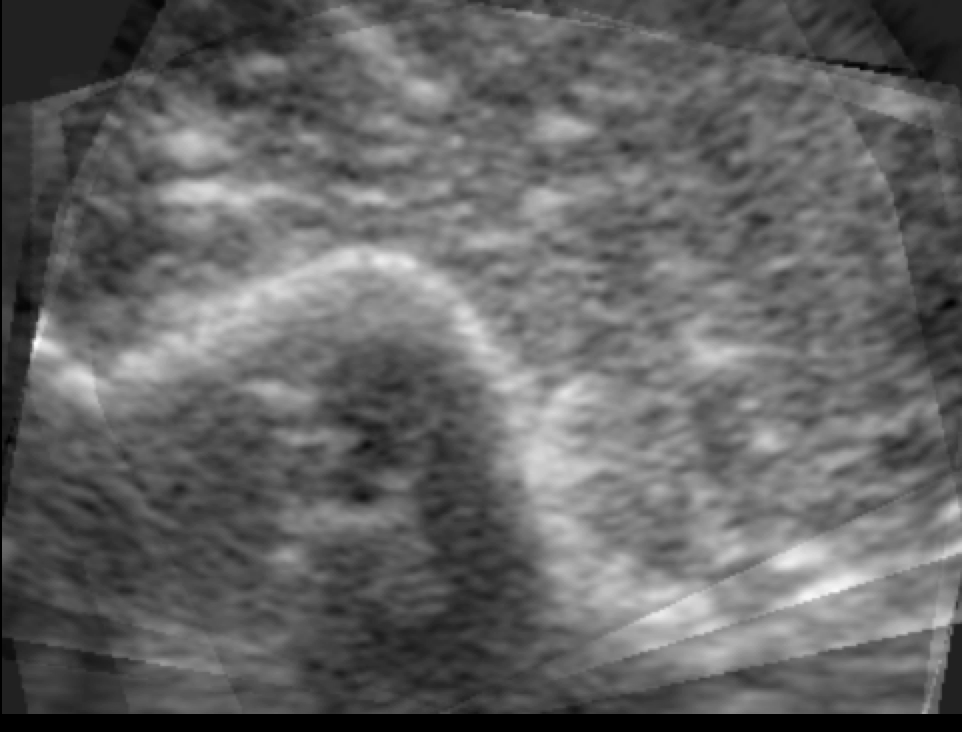}
  \includegraphics[width=0.24\textwidth]{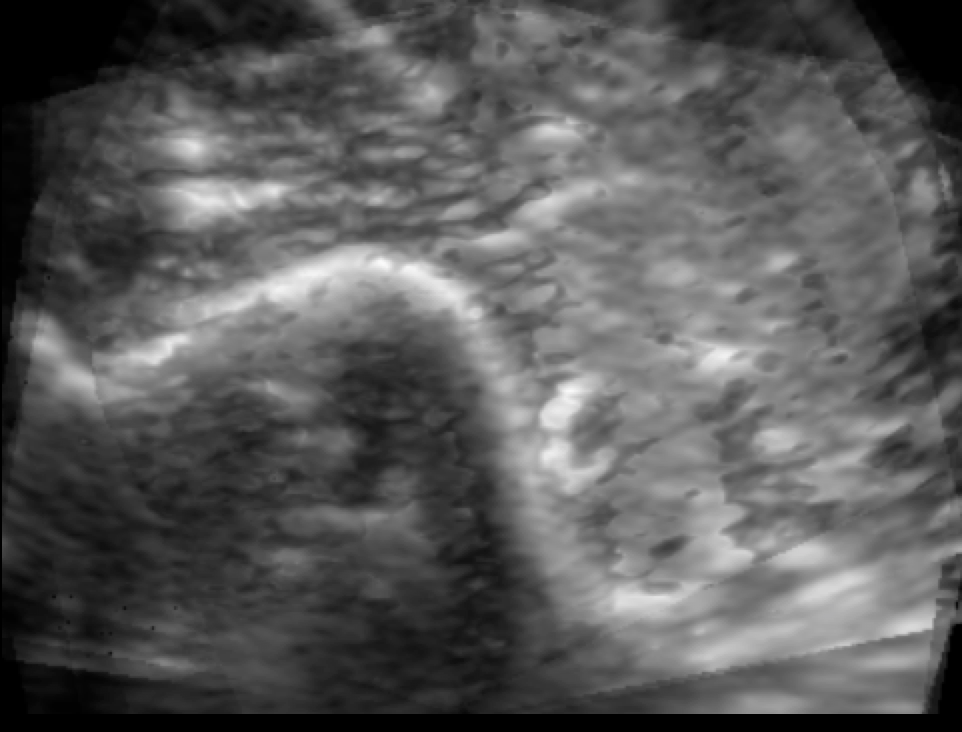}
  \includegraphics[width=0.24\textwidth]{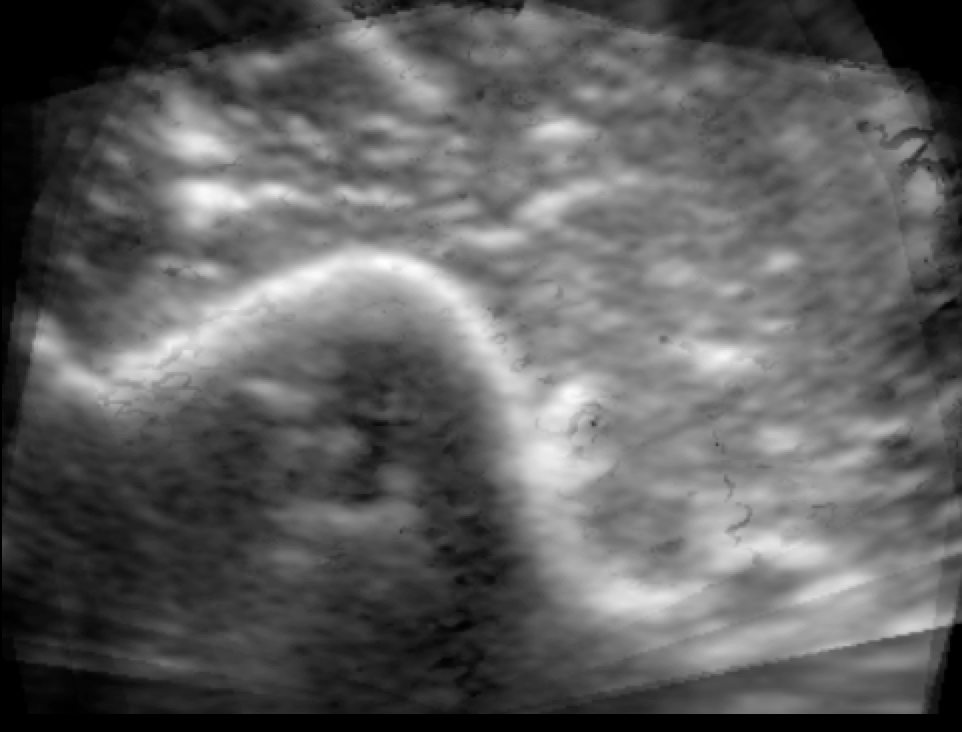}
  \caption{Dataset 1 - 5th image out of a sequence of nine images}
  \label{fig:sfig15}
\end{subfigure}

\begin{subfigure}{0.99\textwidth}
  \centering
  \includegraphics[width=0.24\textwidth]{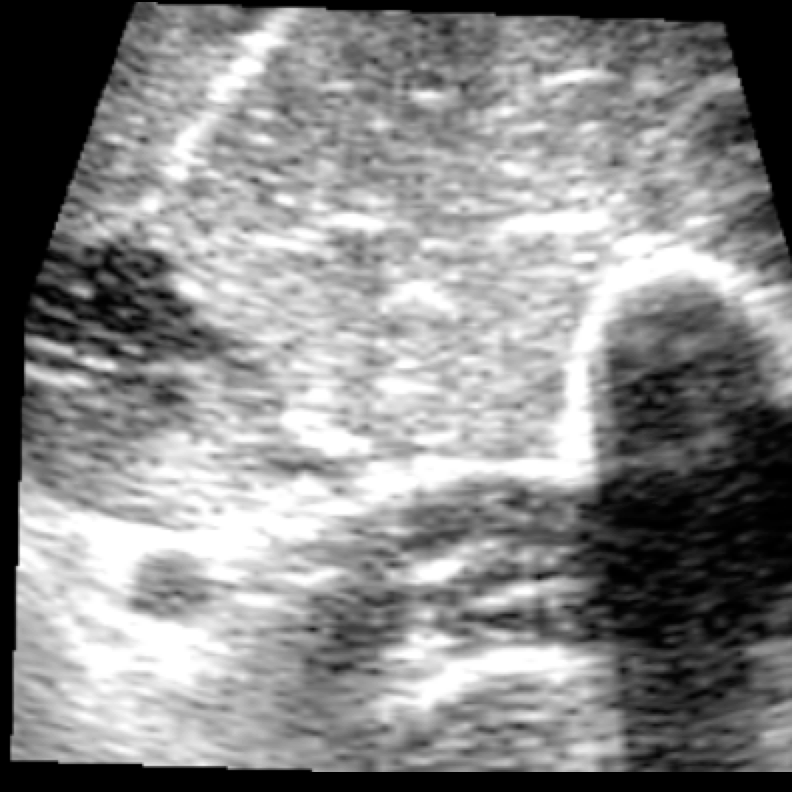}
  \includegraphics[width=0.24\textwidth]{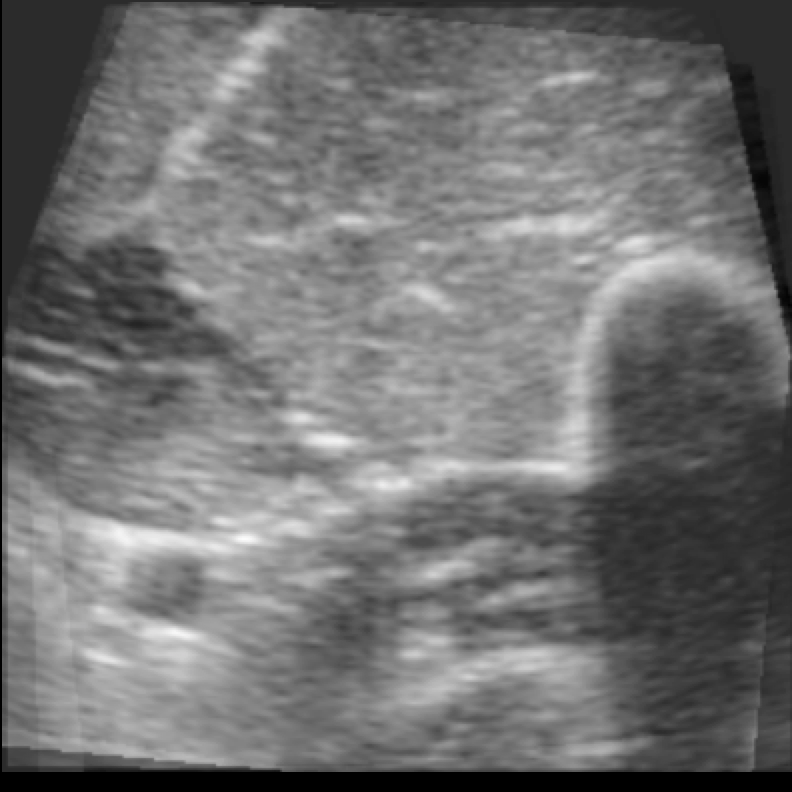}
  \includegraphics[width=0.24\textwidth]{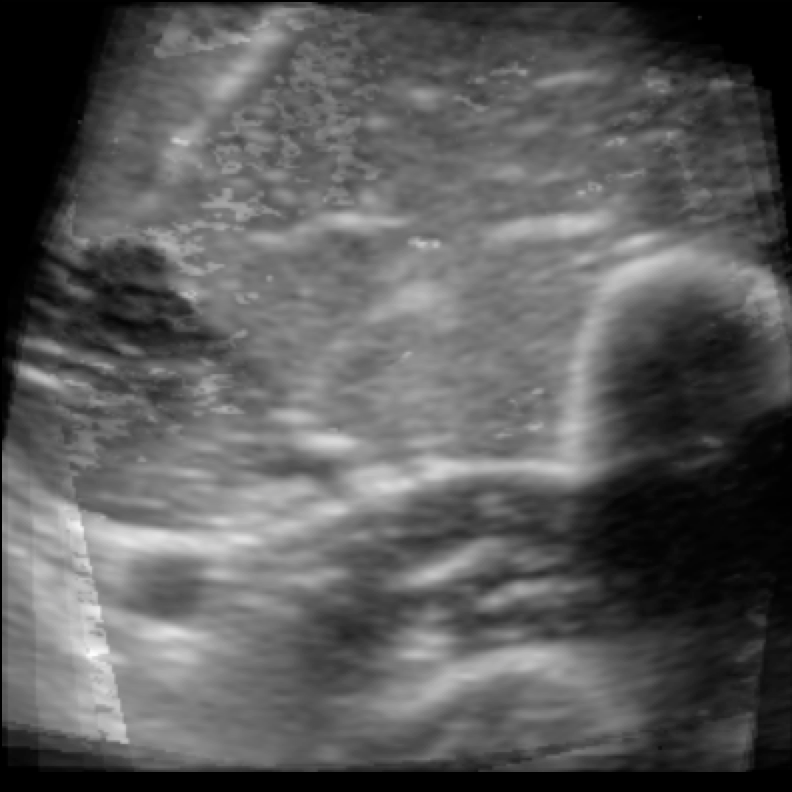}
  \includegraphics[width=0.24\textwidth]{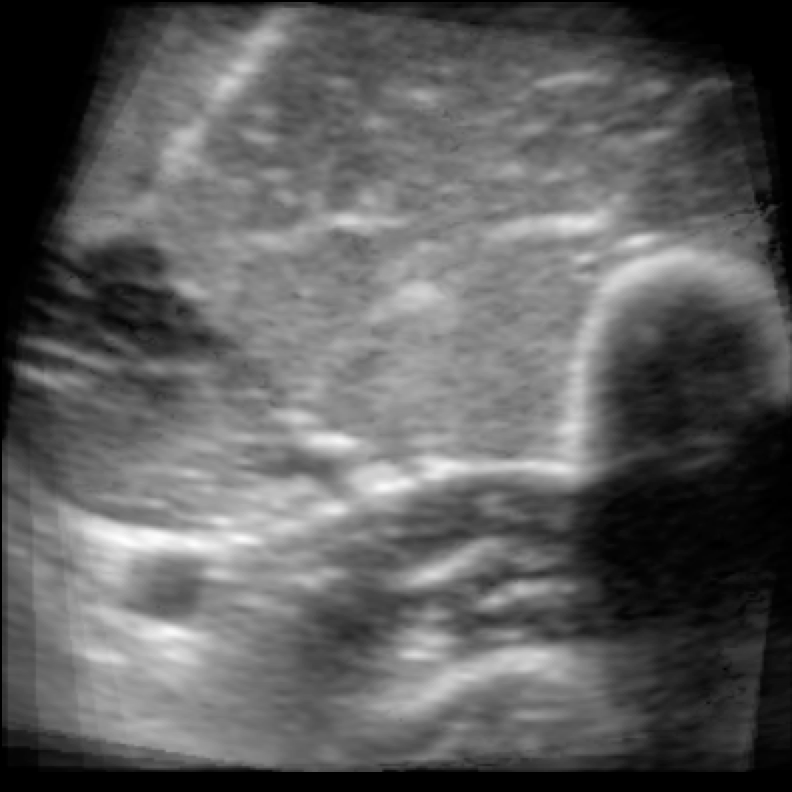}
  \caption{Dataset 2 - 6th image out of a sequence of nine images}
  \label{fig:sfig26}
\end{subfigure}

\begin{subfigure}{0.99\textwidth}
  \centering
  \includegraphics[width=0.24\textwidth]{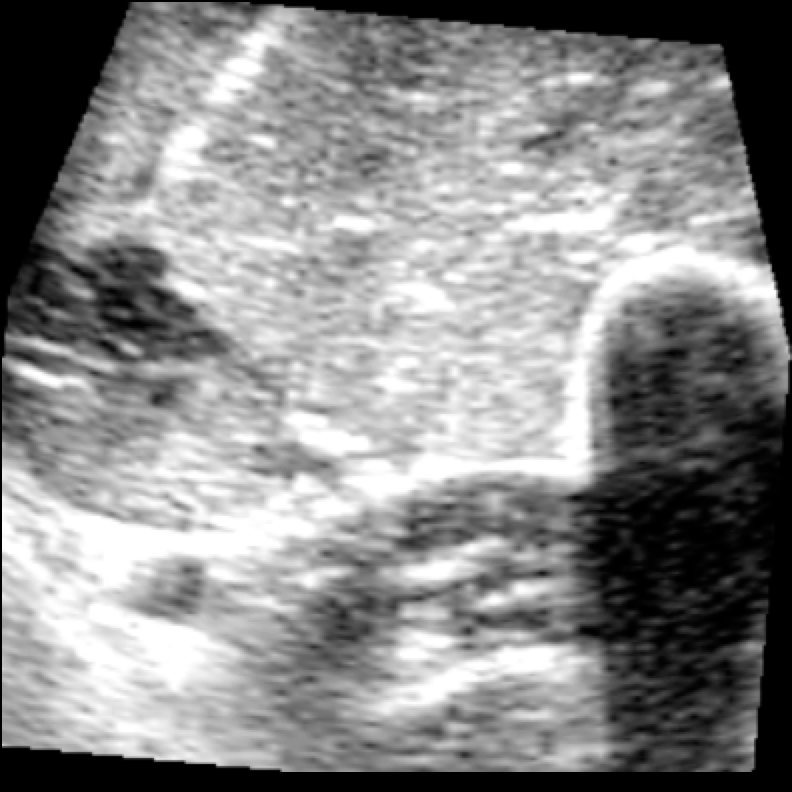}
  \includegraphics[width=0.24\textwidth]{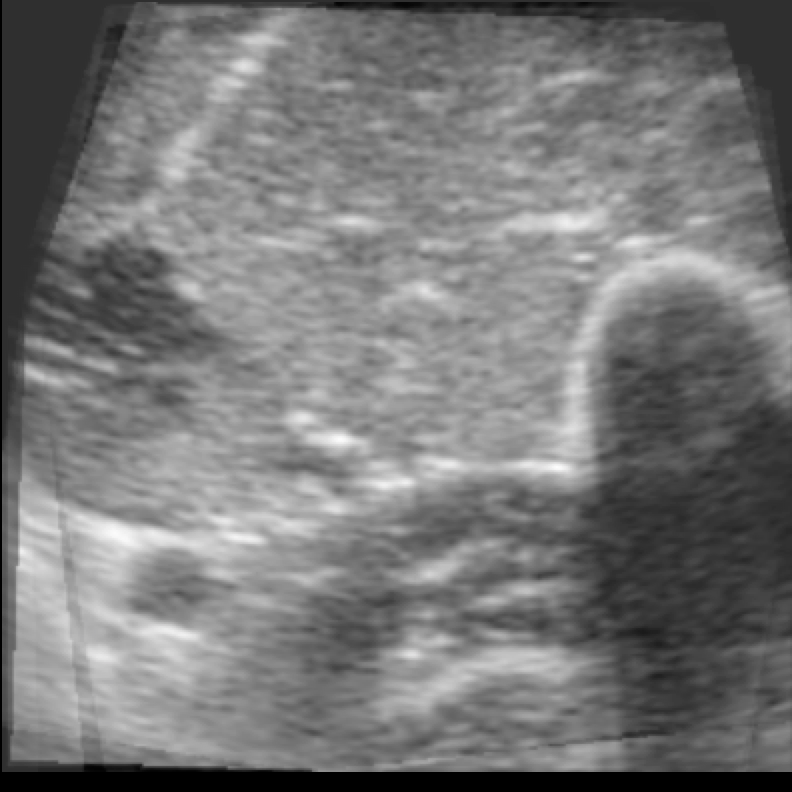}
  \includegraphics[width=0.24\textwidth]{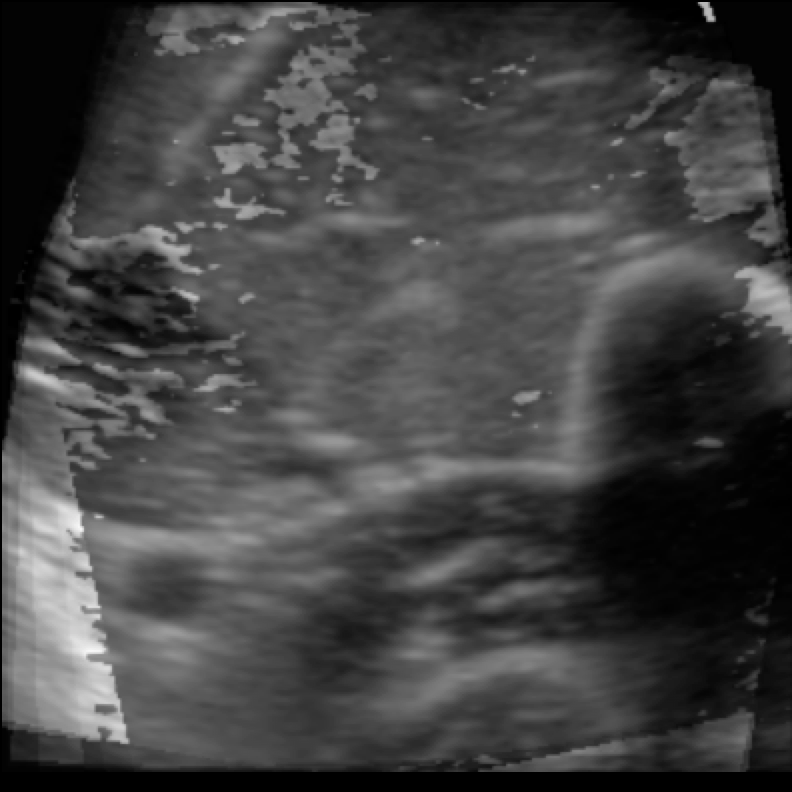}
  \includegraphics[width=0.24\textwidth]{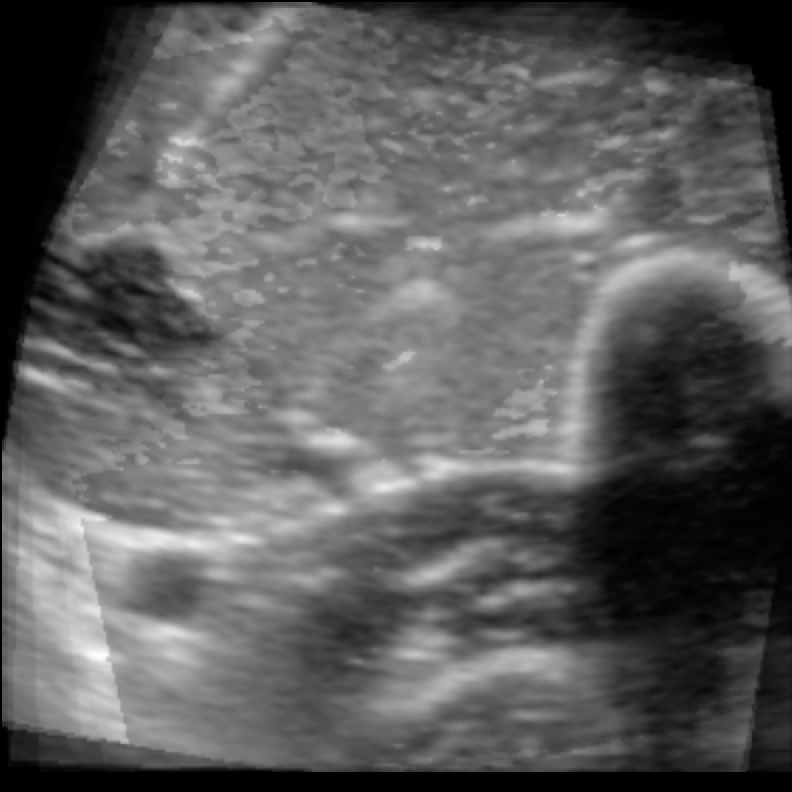}
  \caption{Dataset 2 - 7th image out of a sequence of nine images}
  \label{fig:sfig27}
\end{subfigure}

\caption{The four column from left to right are as follows: (1)
  Leave-one-out image, (2) Hennersperger et
  al.\cite{DBLP:conf/miccai/HennerspergerBM15}, (3) Our method
  $\lambda$ = 0 and, (4) Our method $\lambda$ = 10.} 
\label{fig:fig}
\end{figure}

\section{Results}\label{sec:Experiments and Discussion}
We evaluated the proposed method on in vivo human datasets. The
ultrasound datasets 
were acquired from two volunteers. Consent was obtained before
ultrasound acquisition.
Ultrasound acquisition was performed using a Voluson E10 ultrasound
system with an eM6C curved matrix electronic 4D transducer (GE
Healthcare, Chicago, Il).
Each dataset contained nine (N =
9) volumes.
The probe was gradually 
translated over the body surface whilst tilting to maintain the target
body part in the field of view.   
We use peak signal to noise ratio (PSNR) as the evaluation metric. 

\begin{table}[htb]
\caption{PSNR Leave-one-out results (in dB).} 
\centering 
\begin{tabular}{cccccc} 
\hline\hline 
Dataset& \text{Hennersperger et al.\cite{DBLP:conf/miccai/HennerspergerBM15}} & \multicolumn{4}{c}{Our method} \\
&  & $\lambda$ = 0 & $\lambda$ = 1 & $\lambda$ = 10 & $\lambda$ = 100  \\ [0.5ex]
\hline 
1 & 18.8 & 16.1 & 21.2 & 21.6 &13.6 \\ 
2 & 17.5 & 17.7 & 21.8 & 22.6 & 22.9 \\[0.5ex]
\hline 
\end{tabular}
\label{tab:LooRounds}
\end{table}

The two parameters in \eqref{eq:energy} are $\lambda$ and
$\delta$. $\lambda$ is the scale factor and $\delta$ is the constant
in the Huber loss function. 
$\lambda$ is evaluated for the following set of values: 0, 1, 10 and
100. The constant $\delta$ was set to a small positive value ($\delta
= 0.01$). 

We used leave-one-out strategy to evaluate the performance of the
method. In each round of the leave-one-out we leave out one of the
images from the set of N images. We then estimate the tensor image
using the rest of the N-1 images. The estimated tensor image is used
to generate the projection image along the direction of the left out
image. The projection image is then compared to the left out image
using the PSNR metric. This is iterated over all the images in the
set. The results are averaged over all the round to estimate the
overall error, see
Table~\ref{tab:LooRounds}. Table~\ref{tab:LooRounds} shows that our
method performs better than Hennersperger et
al.\cite{DBLP:conf/miccai/HennerspergerBM15}. As per the
leave-one-out rounds in Table~\ref{tab:LooRounds}, $\lambda$ = 10 is
the best parameter setting. Some of the leave-one-out results are
shown in Figure~\ref{fig:fig}. In Figure~\ref{fig:fig}, the images
on the left columns are the leave-one-out images, the second to left
column are the output images using Hennersperger et
al.\cite{DBLP:conf/miccai/HennerspergerBM15}, the third to left
column are the output images using our method with parameter $\lambda$
= 0, and the last column at the right are the output images using our
method with parameter $\lambda$ = 10. 

\section{Conclusion}
%
We propose a spatial compounding technique to improve the
3D ultrasound image quality by compositing multiple ultrasound volumes
acquired from different probe orientations. Our compounding technique
uses a tensor representation which 
is sensitive to the probe orientation. The proposed method has a
better PSNR than Hennersperger et
al.\cite{DBLP:conf/miccai/HennerspergerBM15} which uses similar tensor
based representation. The log-Euclidean framework ensures that the
tensors are positive definite, enforcing a non-negative image. The
additional total variation term is used for spacial
regularisation. The initial results of the proposed method are
promising. Future work will focus on improving the validation of our
methods, on introducing more realistic models of signal attenuation
and on providing a combined method to jointly optimise the
image alignment and the tensor model fitting.

\acknowledgments
This work was supported by Wellcome / Engineering and Physical
Sciences Research Council (EPSRC) [WT101957; NS/A000027/1;
203145Z/16/Z; NS/A000050/1]

\bibliographystyle{spiebib}
\bibliography{2017-SPIE-ComputationalSonography}

\end{document}